\documentclass{article}

\usepackage[preprint]{nips_2018}

\usepackage[utf8]{inputenc} % allow utf-8 input
\usepackage[T1]{fontenc}    % use 8-bit T1 fonts
\usepackage{hyperref}       % hyperlinks
\usepackage{url}            % simple URL typesetting
\usepackage{booktabs}       % professional-quality tables
\usepackage{amsfonts}       % blackboard math symbols
\usepackage{nicefrac}       % compact symbols for 1/2, etc.
\usepackage{microtype}      % microtypography
\usepackage{graphicx}
\usepackage{siunitx}
\usepackage{eqnarray,amsmath}
\usepackage{xcolor}
\usepackage{bm}
\usepackage{caption}
\usepackage{subcaption}

\def\th{\theta}
\def\vet{\eta}

\newcommand{\dd}[2]{\frac{\text{d} #1}{\text{d} #2}}
\newcommand{\ddinline}[2]{{\text{d} #1}/{\text{d} #2}}
\newcommand{\pp}[2]{\frac{\partial{#1}}{\partial{#2}}}
\newcommand{\ppinline}[2]{{\partial{#1}}/{\partial{#2}}}
\newcommand{\ppp}[3]{\frac{\partial^2{#1}}{\partial{#2}\  \partial{#3}}}

\title{Meta-Gradient Reinforcement Learning}

\author{
  Zhongwen Xu \\
  DeepMind\\
  \texttt{zhongwen@google.com} \\
  \And
  Hado van Hasselt \\
  DeepMind\\
  \texttt{hado@google.com} \\
  \And
  David Silver\\
  DeepMind\\
  \texttt{davidsilver@google.com}  %% Coauthor \\
}

\begin{document}

\maketitle

\begin{abstract}
The goal of reinforcement learning algorithms is to estimate and/or optimise the value function. 
However, unlike supervised learning, no teacher or oracle is available to provide the true value function. Instead, the majority of reinforcement learning algorithms estimate and/or optimise a proxy for the value function. This proxy is typically based on a sampled and bootstrapped approximation to the true value function, known as a \emph{return}. The particular choice of return is one of the chief components determining the nature of the algorithm: the rate at which future rewards are discounted; when and how values should be bootstrapped; or even the nature of the rewards themselves. It is well-known that these decisions are crucial to the overall success of RL algorithms. We discuss a gradient-based meta-learning algorithm that is able to adapt the nature of the return, online, whilst interacting and learning from the environment. When applied to 57 games on the Atari 2600 environment over 200 million frames, our algorithm achieved a new state-of-the-art performance.
\end{abstract}

The central goal of reinforcement learning (RL) is to optimise the agent's \emph{return} (cumulative reward); this is typically achieved by a combination of prediction and control. The prediction subtask is to estimate the value function -- the expected return from any given state. Ideally, this would be achieved by updating an approximate value function towards the true value function. The control subtask is to optimise the agent's policy for selecting actions, so as to maximise the value function. Ideally, the policy would simply be updated in the direction that increases the true value function. However, the true value function is unknown and therefore, for both prediction and control, a sampled return is instead used as a proxy. A large family of RL algorithms \citep{sutton1988learning, Rummery:1994, Seijen:2009, Sutton:2018}, including several state-of-the-art deep RL algorithms \citep{mnih2015human, van2016deep, Harutyunyan:2016, hessel2017rainbow, espeholt2018impala}, are characterised by different choices of the return.  

The \emph{discount factor} $\gamma$ determines the time-scale of the return. 
A discount factor close to $\gamma=1$ provides a long-sighted goal that accumulates rewards far into the future, while a discount factor close to $\gamma=0$ provides a short-sighted goal that prioritises short-term rewards. Even in problems where long-sightedness is clearly desired, it is frequently observed that discounts $\gamma < 1$ achieve better results \citep{Prokhorov:1997}, especially during early learning. It is known that many algorithms converge faster with lower discounts \citep{Bertsekas:1996}, but of course too low a discount can lead to highly sub-optimal policies that are too myopic. In practice it can be better to first optimise for a myopic horizon, e.g., with $\gamma=0$ at first, and then to repeatedly increase the discount only after learning is somewhat successful~\citep{Prokhorov:1997}.

The return may also be \emph{bootstrapped} at different time horizons. An \emph{$n$-step return} accumulates rewards over $n$ time-steps and then adds the value function at the $n$th time-step. The $\lambda$-return \citep{sutton1988learning,Sutton:2018} is a geometrically weighted combination of $n$-step returns. In either case, the meta-parameter $n$ or $\lambda$ can be important to the performance of the algorithm, trading off bias and variance. Many researchers have sought to automate the selection of these parameters \citep{kearns2000bias,downey2010temporal,konidaris2011td_gamma,white2016greedy}. 

There are potentially many other design choices that may be represented in the return, including off-policy corrections \citep{espeholt2018impala, retrace}, target networks \citep{mnih2015human}, emphasis on certain states \citep{emphatic}, reward clipping \citep{mnih2013playing}, or even the nature of the rewards themselves \citep{randlov1998learning, Singh:2005, zheng2018learning}.

In this work, we are interested in one of the fundamental problems in reinforcement learning: what would be the best form of return for the agent to maximise? Specifically, we propose to \emph{learn} the return function by treating it as a parametric function with tunable meta-parameters $\vet$, for instance including the discount factor $\gamma$, or the bootstrapping parameter $\lambda$~\citep{sutton1988learning}. The meta-parameters $\vet$ are adjusted \emph{online} during the agent's interaction with the environment, allowing the return to both adapt to the specific problem, and also to dynamically adapt over time to the changing context of learning. We derive a practical gradient-based meta-learning algorithm and show that this can significantly improve performance on large-scale deep reinforcement learning applications.

\section{Meta-Gradient Reinforcement Learning Algorithms}

In deep reinforcement learning, the value function and policy are approximated by a neural network with parameters $\th$, denoted by $v_{\th}(S)$ and $\pi_{\th}(A|S)$ respectively. At the core of the algorithm is an \emph{update function}, 
\begin{equation}
\th' = \th + f(\tau, \th, \vet) \,,
\end{equation}
that adjusts parameters from a sequence of experience $\tau_t = \{S_t, A_t, R_{t+1}, \ldots \}$ consisting of states $S$, actions $A$ and rewards $R$. The nature of the function is determined by \emph{meta-parameters} $\vet$. 

Our meta-gradient RL approach is based on the principle of online cross-validation~\citep{sutton1992adapting}, using successive samples of experience. The underlying RL algorithm is applied to the first sample (or samples), and its performance is measured in a subsequent sample. Specifically, the algorithm starts with parameters $\th$, and applies the update function to the first sample(s), resulting in new parameters $\th'$; the gradient $\ddinline{\th'}{\vet}$ of these updates indicates how the meta-parameters affected these new parameters.  The algorithm then measures the performance of the new parameters $\th'$ on a subsequent, independent sample $\tau'$, utilising a differentiable \emph{meta-objective} $J'(\tau', \th', \vet')$. When validating the performance on the second sample, we use a fixed meta-parameter $\vet'$ in $J'$ as a reference value. In this way, we form a differentiable function of the meta-parameters, and obtain the gradient of $\vet$ by taking the derivative of meta-objective $J'$ w.r.t. $\vet$ and applying the chain rule:
\begin{equation}
\label{eq:meta_chain_rule}
\pp{J'(\tau', \th', \vet')}{\vet} = \pp{J'(\tau', \th', \vet')}{\th'} \dd{\th'}{\vet} \,.
\end{equation}
To compute the gradient of the updates, $\text{d}\th'/\text{d}\vet$, we note that the parameters form an additive sequence, and the gradient can therefore be accumulated online,
\begin{align}
\label{eq:meta_gradient}
\dd{\th'}{\vet} &= \dd{\th}{\vet} + \pp{f(\tau, \th, \vet)}{\vet} + \pp{f(\tau, \th, \vet)}{\th} \dd{\th}{\vet} = \left(I + \pp{f(\tau, \th, \vet)}{\th}\right) \dd{\th}{\vet} + \pp{f(\tau, \th, \vet)}{\vet}
\end{align}
The gradient $\ppinline{f(\tau, \th, \vet)}{\th}$ is large and challenging to compute in practice. Instead we approximate the gradient using an accumulative trace $z \approx \ddinline{\th}{\vet}$,
\begin{align}
\label{eq:meta_trace}
z' &= \mu z + \pp{f(\tau, \th, \vet)}{\vet}
\end{align}
The gradient in Equation (\ref{eq:meta_gradient}) is defined for fixed meta-parameters $\vet$. In practice, the meta-parameters will adapt online. To allow for this adaptation, the parameter $\mu \in [0,1]$ decays the trace and focuses on recent updates. Choosing $\mu=0$ results in a greedy meta-gradient that considers only the immediate effect of the meta-parameters $\vet$ on the parameters $\theta$; this may often be sufficient.

Finally, the meta-parameters $\vet$ are updated to optimise the meta-objective, for example by applying stochastic gradient descent (SGD) to update $\vet$ in the direction of the meta-gradient,
\begin{equation}
\label{eq:update_eta}
\Delta \vet = -\beta \pp{J'(\tau', \th', \vet')}{\th'} z', 
\end{equation}
where $\beta$ is the learning rate for updating meta parameter $\vet$.

In the following sections we instantiate this idea more specifically to RL algorithms based on predicting or controlling returns. We begin with a pedagogical example of using meta-gradients for prediction using a temporal-difference update. We then consider meta-gradients for control, using a canonical actor-critic update function and a policy gradient meta-objective. Many other instantiations of meta-gradient RL would be possible, since the majority of deep reinforcement learning updates are differentiable functions of the return, including, for instance, value-based methods like SARSA($\lambda$) \citep{Rummery:1994, Sutton:2018} and DQN \citep{mnih2015human}, policy-gradient methods \citep{Williams:1992}, or actor-critic algorithms like A3C~\citep{mnih2016asynchronous} and IMPALA~\citep{espeholt2018impala}.  

\subsection{Applying Meta-Gradients to Returns}

We define the return $g_{\vet}(\tau_t)$ to be a function of an episode or a truncated $n$-step sequence of experience $\tau_t = \{S_t, A_t, R_{t+1}, \ldots, S_{t+n} \}$. The nature of the return is determined by the meta-parameters $\vet$. 

The $n$-step return \citep{Sutton:2018} accumulates rewards over the sequence and then bootstraps from the value function,
\begin{equation}
g_{\vet}(\tau_t) = R_{t+1} + \gamma R_{t+2} + \gamma^2 R_{t+3} + \ldots, + \gamma^{n-1} R_{t+n} + \gamma^{n} v_{\th}(S_{t+n}) \,
\end{equation}
where $\vet = \{\gamma, n\}$.

The $\lambda$-return is a geometric mixture of $n$-step returns,
\citep{sutton1988learning}
\begin{equation}
g_{\vet}(\tau_t) = R_{t+1} + \gamma (1 - \lambda) v_{\th}(S_{t+1}) + \gamma \lambda g_{\vet}(\tau_{t+1}) \,
\end{equation}
where $\vet = \{\gamma, \lambda\}$. The $\lambda$-return has the advantage of being fully differentiable with respect to the meta-parameters. The meta-parameters $\vet$ may be viewed as gates that cause the return to terminate ($\gamma=0$) or bootstrap $(\lambda=0)$, or to continue onto the next step ($\gamma=1$ and $\lambda=1$). The $n$-step or $\lambda$-return can be augmented with off-policy corrections ~\citep{Precup:2000, Sutton:2014, espeholt2018impala} if it is necessary to correct for the distribution used to generate the data.

A typical RL algorithm would hand-select the meta-parameters, such as the discount factor $\gamma$ and bootstrapping parameter $\lambda$, and these would be held fixed throughout training. Instead, we view the return $g$ as a function parameterised by meta-parameters $\vet$, which may be differentiated to understand its dependence on $\vet$. This in turn allows us to compute the gradient $\ppinline{f}{\vet}$ of the update function with respect to the meta-parameters $\vet$, and hence the meta-gradient $\ppinline{J'(\tau', \th', \vet')}{\vet}$. In essence, our agent asks itself the question, ``which return results in the best performance?", and adjusts its meta-parameters accordingly. 

\subsection{Meta-Gradient Prediction}
\label{sec:prediction}

We begin with a simple instantiation of the idea, based on the canonical TD($\lambda$) algorithm for prediction. The objective of the TD($\lambda$) algorithm (according to the forward view \citep{Sutton:2018}) is to minimise the  squared error between the value function approximator $v_{\th}(S)$ and the $\lambda$-return $g_\vet(\tau)$, 
\begin{align}
J(\tau, \th, \vet) &= (g_{\vet}(\tau) - v_{\th}(S))^2 &&&
\pp{J(\tau,\th, \vet)}{\th} &= -2 (g_{\vet}(\tau) - v_{\th}(S)) \pp{v_{\th}(S)}{\th}
\label{eq:tdlambda}
\end{align}
where $\tau$ is a sampled trajectory starting with state $S$, and $\ppinline{J(\tau,\th, \vet)}{\th}$ is a semi-gradient ~\citep{Sutton:2018}, i.e. the $\lambda$-return is treated as constant with respect to $\th$.

The TD($\lambda$) update function $f(\cdot)$ applies SGD to update the agent's parameters $\th$ to descend the gradient of the objective with respect to the parameters,
\begin{align}
f(\tau,\th,\vet) &= - \frac{\alpha}{2} \pp{J(\tau,\th,\vet)}{\th} =  \alpha (g_{\vet}(\tau) - v_{\th}(S)) \pp{v_{\th}(S)}{\th}
\label{eq:sgd}
\end{align}
where $\alpha$ is the learning rate for updating agent $\th$. We note that this update is itself a differentiable function of the meta-parameters $\vet$,
\begin{align}
\pp{f(\tau,\th,\vet)}{\vet} = - \frac{\alpha}{2} \ppp{J(\tau,\th,\vet)}{\th}{\vet} = \alpha \pp{g_{\vet}(\tau)}{\vet} \pp{v_{\th}(S)}{\th}
\label{eq:tdlambda_meta}
\end{align}
The key idea of the meta-gradient prediction algorithm is to adjust meta-parameters $\vet$ in the direction that achieves the best predictive accuracy. This is measured by cross-validating the new parameters $\th'$ on a second trajectory $\tau'$ that starts from state $S'$, using a mean squared error (MSE) meta-objective and taking its semi-gradient,
\begin{align}
J'(\tau', \th', \vet') &= (g_{\vet'}(\tau') - v_{\th'}(S'))^2 &&&
\pp{J'(\tau',\th', \vet')}{\th'} &= -2 (g_{\vet'}(\tau') - v_{\th'}(S')) \pp{v_{\th'}(S')}{\th'}
\label{eq:tdlambda_meta_loss}
\end{align}
The meta-objective in this case could make use of an unbiased and long-sighted return\footnote{The meta-objective could even use a discount factor that is longer-sighted than the original problem, perhaps spanning over many episodes.}, for example using $\vet' = \{\gamma',\lambda'\}$ where $\gamma'=1$ and $\lambda'=1$.

\subsection{Meta-Gradient Control}

We now provide a practical example of meta-gradients applied to control. We focus on the A2C algorithm -- an actor-critic update function that combines both prediction and control into a single update. This update function is widely used in several state-of-the-art agents \citep{mnih2016asynchronous,jaderberg2016reinforcement,espeholt2018impala}. The semi-gradient of the A2C objective, $\ppinline{J({\tau}; \th, \vet)}{\th}$, is defined as follows,
\begin{equation}
-\pp{J(\tau,\th,\vet)}{\th} = (g_{\vet}({\tau}) - v_{\th}(S)) \pp{\log\pi_{\th}(A | S)}{\th} + b (g_{\vet}(\tau) - v_{\th}(S)) \pp{v_{\th}(S)}{\th} + c \pp{H(\pi_{\th}(\cdot|S))}{\th}.
\label{eq:a2c_loss}
\end{equation}

The first term represents a control objective, encouraging the policy $\pi_{\th}$ to select actions that maximise the return. The second term represents a prediction objective, encouraging the value function approximator $v_{\th}$ to more accurately estimate the return $g_{\vet}(\tau)$. The third term regularises the policy according to its entropy $H(\pi_{\th})$, and $b$, $c$ are scalar coefficients that weight the different components in the objective function. 

The A2C update function $f(\cdot)$ applies SGD to update the agent's parameters $\th$. This update function is a differentiable function of the meta-parameters $\vet$, 
\begin{align}
f(\tau,\th,\vet) &= -\alpha \pp{J(\tau,\th,\vet)}{\th} &&&
\pp{f(\tau,\th,\vet)}{\vet} &= \alpha \pp{g_{\vet}(\tau)}{\vet} \left[ \pp{\log \pi_{\th}(A|S)}{\th} + c  \pp{v_{\th}(S)}{\th} \right] \label{eq:meta_final}
\end{align}
Now we come to the choice of meta-objective $J'$ to use for control. Our goal is to identify the return function that maximises overall performance in our agents. This may be directly measured by a meta-objective focused exclusively on optimising returns -- in other words a policy gradient objective, 
\begin{equation} \label{eq:meta_loss}
\pp{J'(\tau', \th', \vet')}{\th'} = (g_{\vet'}(\tau') - v_{\th'} (S')) \pp{\log \pi_{\th'} (A'|S')}{\th'}.
\end{equation}

This equation evaluates how good the updated policy $\th'$ is in terms of returns computed under $\vet'$, when measured on ``held-out'' experiences $\tau'$, e.g. the subsequent $n$-step trajectory. When cross-validating performance using this meta-objective, we use fixed meta-parameters $\vet'$, ideally representing a good proxy to the true objective of the agent. In practice this typically means selecting reasonable values of $\vet'$; the agent is free to adapt its meta-parameters $\vet$ and choose values that perform better in practice. 

We now put the meta-gradient control algorithm together. First, the parameters $\th$ are updated on a sample of experience $\tau$ using the A2C update function (Equation \eqref{eq:a2c_loss}), and the gradient of the update (Equation \eqref{eq:meta_final}) is accumulated into trace $z$. Second, the performance is cross-validated on a subsequent sample of experience $\tau'$ using the policy gradient meta-objective (Equation \eqref{eq:meta_loss}). Finally, the meta-parameters $\vet$ are updated according to the gradient of the meta-objective (Equation \eqref{eq:update_eta}).

\subsection{Conditioned Value and Policy Functions}

One complication of the approach outlined above is that the return function $g_{\vet}(\tau)$ is non-stationary, adapting along with the meta-parameters throughout the training process. As a result, there is a danger that the value function $v_{\th}$ becomes inaccurate, since it may be approximating old returns. For example, the value function may initially form a good approximation of a short-sighted return with $\gamma=0$, but if $\gamma$ subsequently adapts to $\gamma=1$ then the value function may suddenly find its approximation is rather poor. The same principle applies for the policy $\pi$, which again may have specialised to old returns. 

To deal with non-stationarity in the value function and policy, we utilise an idea similar to universal value function approximation (UVFA)~\citep{schaul2015universal}. The key idea is to provide the meta-parameters $\vet$ as an additional input to condition the value function and policy, as follows:
\begin{align*}
v_{\th}^{\vet}(S) &= v_{\th}([S ; {\bf e}_{\vet}]), &&&
\pi_{\th}^{\vet}(S) &= \pi_{\th}([S ; {\bf e}_{\vet}]), &&&
{\bf e}_{\vet} &= \mathbf{W}_{\vet} \vet,
\end{align*}
where ${\bf e}_{\vet}$ is the embedding of $\vet$, $[s; {\bf e}_{\vet}]$ denotes concatenation of vectors $s$ and ${\bf e}_{\vet}$, $\mathbf{W}_{\vet}$ is the embedding matrix (or row vector, for scalar $\eta$) that is updated by backpropagation during training. 

In this way, the agent explicitly learns value functions and policies that are appropriate for various $\vet$. The approximation problem becomes a little harder, but the payoff is that the algorithm can freely shift the meta-parameters without needing to wait for the approximator to ``catch up". 

\subsection{Meta-Gradient Reinforcement Learning in Practice}

To scale up the meta-gradient approach, several additional steps were taken. For efficiency, the A2C objective and meta-objective were accumulated over all time-steps within an $n$-step trajectory of experience. The A2C objective was optimised by RMSProp~\citep{tieleman2012lecture} without momentum~\citep{mnih2015human,mnih2016asynchronous,espeholt2018impala}. This is a differentiable function of the meta-parameters, and can therefore be substituted similarly to SGD (see Equation \eqref{eq:meta_final}); this process may be simplified by automatic differentiation (Appendix~\ref{appendix:autodiff}). As in IMPALA, an off-policy correction was used, based on a V-trace return (see Appendix \ref{appendix:vtrace}). For efficient implementation, mini-batches of trajectories were computed in parallel; trajectories were reused twice for both the update function and for cross-validation (see Appendix \ref{appendix:data_efficiency}). 

\section{Illustrative Examples}

\begin{figure}
\begin{subfigure}[t]{.5\textwidth}
    \centering
    \raisebox{0.35\height}{\includegraphics[width=0.9\textwidth]{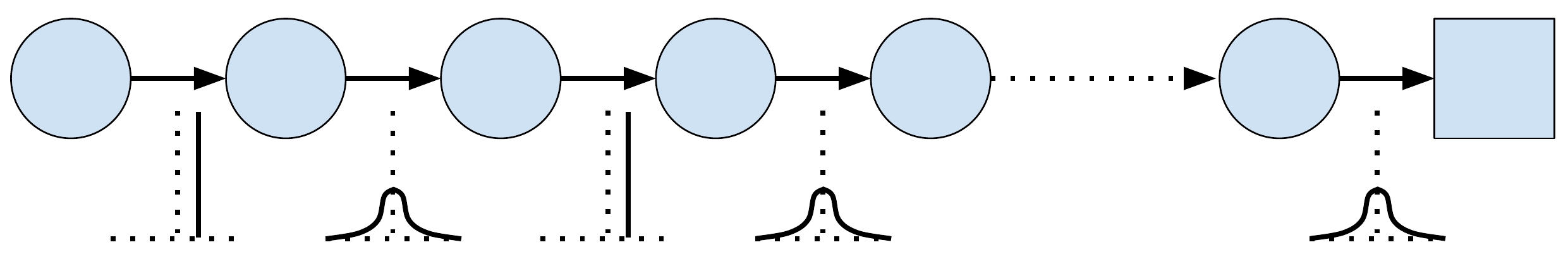}}
    \caption{Rewards alternate between $+0.1$ or randomly sampling from a zero-mean Gaussian.}
\end{subfigure}
\hspace{1ex}
\begin{subfigure}[t]{.5\textwidth}
    \centering
    \includegraphics[width=0.9\textwidth]{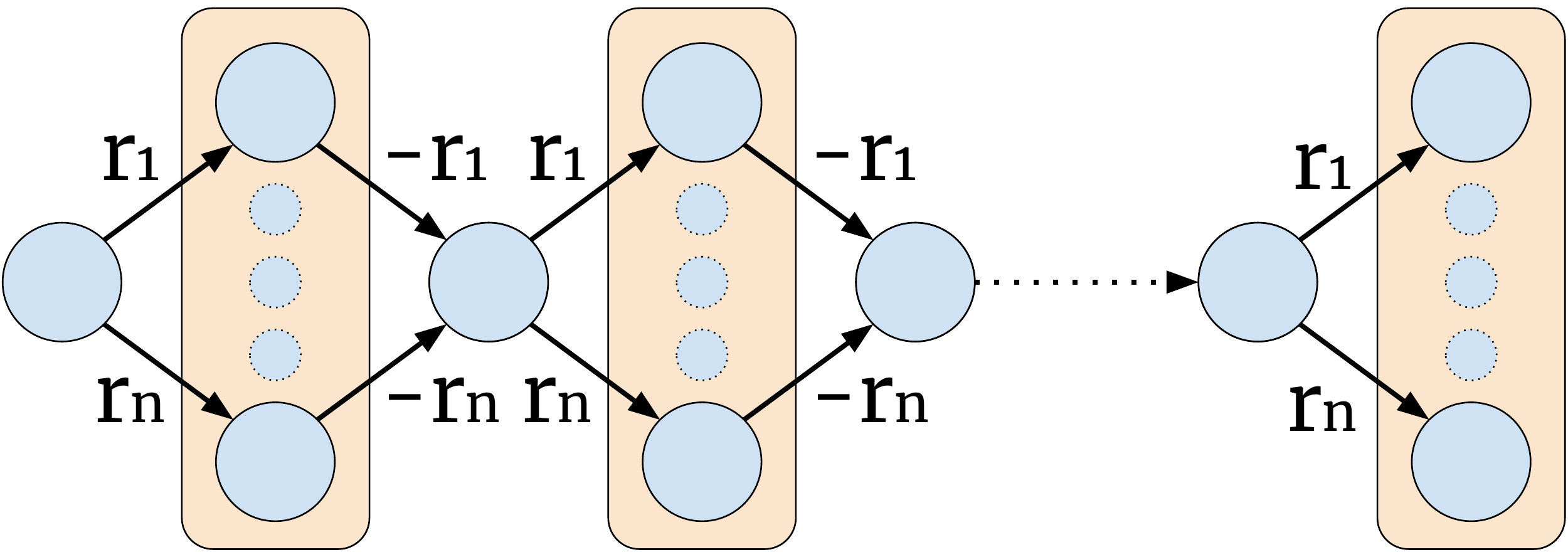}
    \caption{Bottleneck states transition randomly to one of $n$ states; these states (shaded rectangles) are aliased.}
\end{subfigure}
\\[1em]
\begin{subfigure}{.5\textwidth}
    \centering
    \includegraphics[width=1\textwidth]{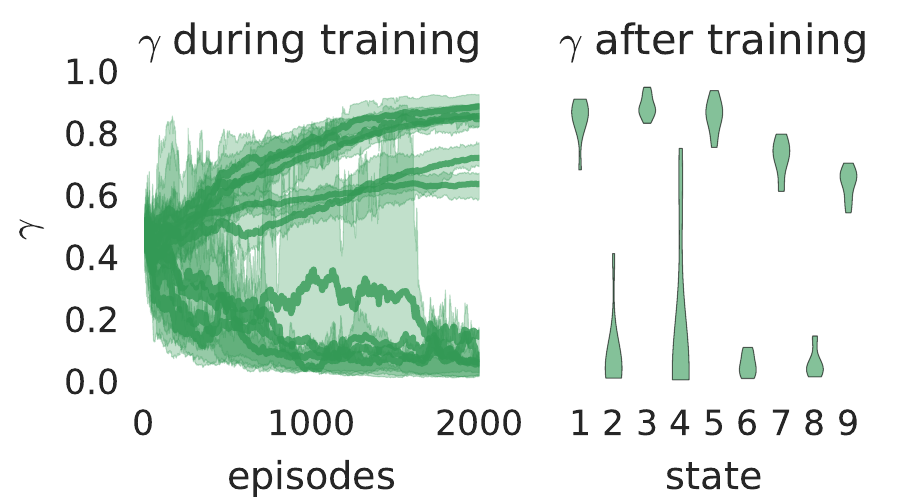}
    % \caption{Rewards alternate between $+0.1$ or randomly sampling from a zero-mean Gaussian}
\end{subfigure}
\hspace{1ex}
\begin{subfigure}{.5\textwidth}
    \centering
    \includegraphics[width=1\textwidth]{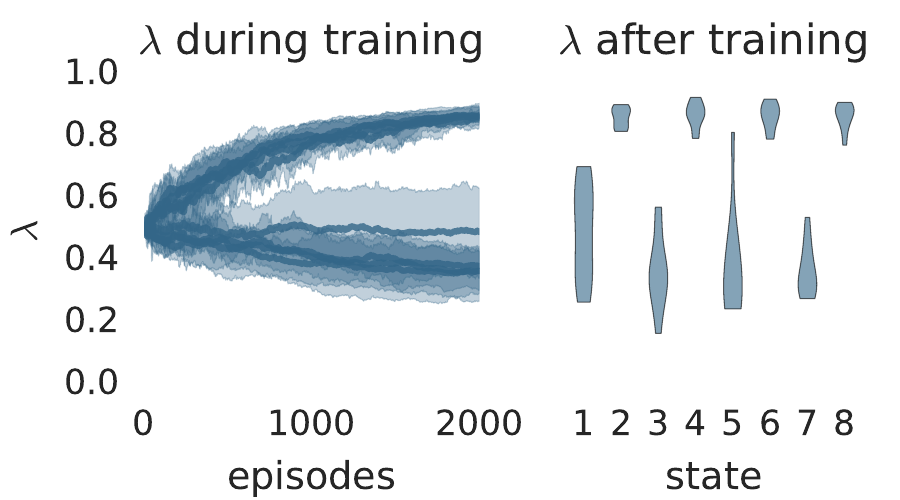}
    % \caption{Bottleneck states transition randomly to one of $n$ states; these states (shaded rectangles) are aliased.}
\end{subfigure}
\caption{Illustrative results of meta-gradient learning of a state-dependent  (a) bootstrapping parameter $\lambda$ or (b) discount factor $\gamma$, in the respective Markov reward processes (top). In each of the subplot shown in the bottom, the first one shows how the meta-parameter $\gamma$ or $\lambda$ adapts over the course of training (averaged over 10 seeds - shaded regions cover 20\%--80\% percentiles). The second plot shows the final value of $\gamma$ or $\lambda$ in each state, identifying appropriately high/low values for odd/even states respectively (violin plots show distribution over seeds). \label{fig:toy}}
\end{figure}

To illustrate the key idea of our meta-gradient approach, we provide two examples that show how the discount factor $\gamma$ and temporal difference parameter $\lambda$, respectively, can be meta-learned. We focus on meta-gradient prediction using the TD($\lambda$) algorithm and a MSE meta-objective with $\gamma'=1$ and $\lambda'=1$, as described in Section \ref{sec:prediction}. For these illustrative examples, we consider \emph{state-dependent} meta-parameters that can take on a different value in each state. 

The first example is a 10-step Markov reward process (MRP), that alternates between ``signal'' and ``noise'' transitions. Transitions from odd-numbered ``signal'' states receive a small positive reward, $R=+0.1$. Transitions from even-numbered ``noise'' states receive a random reward, $R \sim \mathcal{N}(0,1)$. To ensure that the signal can overwhelm the noise, it is beneficial to terminate the return (low $\gamma$) in ``noise'' states, but to continue the return (high $\gamma$) in ``signal'' states. 

The second example is a 9-step MRP, that alternates between ``fan-out'' and ``fan-in'' transitions. At odd-numbered ``fan-out'' time-steps, the state transitions to a randomly sampled successor, all of which are aliased. At even-numbered ``fan-in'' time-steps, the state deterministically transitions back to a bottleneck state. The transition to a ``fan-out'' state yields a deterministic reward unique to that state.  The transition away from that state yields the negation of that same reward. Each pair of ``fan-in'' and ``fan-out" time-steps is zero-mean in expected reward. The fully observed bottleneck states are visited frequently and tend to have more accurate value functions. To predict accurately, it is therefore beneficial to bootstrap (low $\lambda$) in bottleneck states for which the value function is well-known, but to avoid bootstrapping (high $\lambda$) in the noisier, partially observed, fanned-out states.

Figure \ref{fig:toy} shows the results of meta-gradient prediction using the TD($\lambda$) algorithm. The meta-gradient algorithm was able to adapt both $\lambda$ and $\gamma$ to form returns that alternate between high or low values in odd or even states respectively.  

\section{Deep Reinforcement Learning Experiments} 

In this section, we demonstrate the advantages of the proposed meta-gradient learning approach using a state-of-the-art actor-critic framework IMPALA~\citep{espeholt2018impala}. We focused on adapting the discount factor $\vet = \{\gamma\}$ (which we found to be the most effective meta-parameter in preliminary experiments), and the bootstrapping parameter $\lambda$. For these experiments, the meta-parameters were state-independent, adapting one scalar value for $\gamma$ and $\lambda$ respectively (state-dependent meta-parameters did not provide significant benefit in preliminary experiments).\footnote{In practice we parameterise $\vet = \sigma(x)$, where $\sigma$ is the logistic function $\sigma(x) = \frac{1}{1 + e^{-x}}$; i.e. the meta-parameters are actually the logits of $\gamma$ and $\lambda$.}

\subsection{Experiment Setup}

We validate the proposed approach on Atari 2600 video games from Arcade Learning Environment (ALE)~\citep{bellemare2013arcade}, a standard benchmark for deep reinforcement learning algorithms. We build our agent with the IMPALA framework~\citep{espeholt2018impala}, an efficient distributed implementation of actor-critic architecture~\citep{Sutton:2018,mnih2016asynchronous}. We utilise the deep ResNet architecture~\citep{he2016deep} specified in~\citet{espeholt2018impala}, which has shown great advantages over the shallow architecture~\citep{mnih2015human}. Following~\citeauthor{espeholt2018impala}, we train our agent for 200 million frames. Our algorithm does not require extra data compared to the baseline algorithms, as each experience can be utilised in both training the agent itself and training the meta parameters $\vet$ (i.e., each experience can serve as validation data of other experiences). We describe the detailed implementation in the Appendix~\ref{appendix:data_efficiency}. For full details about the IMPALA implementation and the specific off-policy correction $g_{\vet}(\tau)$, please refer to \citet{espeholt2018impala}.

The agents are evaluated on 57 different Atari games and the median of human-normalised scores~\citep{nair2015massively,van2016deep,wang2015dueling,mnih2016asynchronous} are reported. There are two different evaluation protocols. The first protocol is is ``human starts''~\citep{nair2015massively,wang2015dueling, van2016deep}, which initialises episodes to a state that is randomly sampled from human play. The second protocol is ``no-ops starts'', which initialises each episode with a random sequence of no-op actions; this protocol is also used during training. 
We keep all of the hyper-parameters (e.g., batch size, unroll length, learning rate, entropy cost) the same as specified in~\citet{espeholt2018impala} for a fair comparison. For self-contained purpose, we provide all of the important hyper-parameters used in this paper, including the ones following~\citet{espeholt2018impala} and the additional meta-learning optimisation hyper-parameters (i.e., meta batch size, meta learning rate $\alpha'$, embedding size for $\vet$), in Appendix~\ref{appendix:hyperparam}.  The meta-learning hyper-parameters are chosen according to the performance of six Atari games as common practice in Deep RL Atari experiments~\citep{van2016deep, mnih2016asynchronous, wang2015dueling}. Additional implementation details are provided in Appendix~\ref{appendix:impl}.

\subsection{Experiment Results}

We compared four variants of the IMPALA algorithm: the original baseline algorithm without meta-gradients, i.e. $\vet=\{\}$; using meta-gradients with $\vet=\{\lambda\}$; using meta-gradients with $\vet=\{\gamma\}$; and using meta-gradients with $\vet=\{\gamma,\lambda\}$. The original IMPALA algorithm used a discount factor of $\gamma=0.99$; however, when we manually tuned the discount factor and found that a discount factor of $\gamma=0.995$ performed considerably better (see Appendix~\ref{appendix:grid_search_gamma}). For a fair comparison, we tested our meta-gradient algorithm in both cases. When the discount factor is not adapted, $\vet=\{\}$ or $\vet=\{\lambda\}$, we used a fixed value of $\gamma=0.99$ or $\gamma=0.995$. When the discount factor is adapted, $\vet=\{\gamma\}$ or $\vet=\{\gamma,\lambda\}$, we  cross-validate with a meta-parameter of $\gamma'=0.99$ or $\gamma'=0.995$ accordingly in the meta-objective $J'$ (Equation~(\ref{eq:meta_loss})). Manual tuning of the $\lambda$ parameter did not have a significant impact on performance and we therefore compared only to the original value of $\lambda=1$.

We summarise the median human-normalised scores in Table~\ref{tb:combined_results}; individual improvements on each game, compared to the IMPALA baseline, are given in Appendix~\ref{appendix:individual}; and individual plots demonstrating the adaptation of $\gamma$ and $\lambda$ are provided in Appendix~\ref{appendix:training_curves}. The meta-gradient RL algorithm increased the median performance, compared to the baseline algorithm, by a margin between 30\% and 80\% across ``human starts'' and ``no-op starts" conditions, and with both $\gamma=0.99$ and $\gamma=0.995$. 

%%%%
\begin{table}
\centering
\begin{tabular}{lr rr rr}
\toprule
& $\vet$ & \multicolumn{2}{c}{Human starts} & \multicolumn{2}{c}{No-op starts} \\
\midrule
&& $\gamma=0.99$ & $\gamma=0.995$ & $\gamma=0.99$ & $\gamma=0.995$ \vspace{1ex}
\\ 
IMPALA & $\{\}$  & 144.4\% & 211.9\% & 191.8\% & 257.1\% \\
Meta-gradient & $\{\lambda\}$ & 156.6\% & 214.2\% & 185.5\% & 246.5\%  \\
\midrule
&& $\gamma'=0.99$ & $\gamma'=0.995$ & $\gamma'=0.99$ & $\gamma'=0.995$ \vspace{1ex}
\\
Meta-gradient & $\{\gamma\}$ & 233.2\% & 267.9\% & 280.9\% & 275.5\% \\
Meta-gradient & $\{\gamma,\lambda\}$ & 221.6\% & 292.9\% &  242.6\% & 287.6\% \\
\bottomrule
\end{tabular}
\caption{
Results of meta-learning the discount parameter $\gamma$, the temporal-difference learning parameter $\lambda$, or both $\gamma$ and $\lambda$, compared to the baseline IMPALA algorithm which meta-learns neither. Results are given both for the discount factor $\gamma=0.99$ originally reported in \citep{espeholt2018impala} and also for a tuned discount factor $\gamma=0.995$ (see Appendix~\ref{appendix:grid_search_gamma}); the cross-validated discount factor $\gamma'$ in the meta-objective was set to the same value for a fair comparison.
\label{tb:combined_results}
}
\end{table}
%%%%

We also verified the architecture choice of conditioning the value function $v$ and policy $\pi$ on the meta-parameters $\vet$. We compared the proposed algorithm with an identical meta-gradient algorithm that adapts the discount factor $\vet = \{\gamma\}$, but does not provide an embedding of the discount factor as an input to $\pi$ and $v$. For this experiment, we used a cross-validation discount factor of $\gamma' = 0.995$. The human-normalised median score was only 183\%, well below the IMPALA baseline with $\gamma = 0.995$ (211.9\%), and much worse than the full meta-gradient algorithm that includes the discount factor embedding (267.9\%).

Finally, we compare against the state-of-the-art agent trained on Atari games, namely Rainbow~\citep{hessel2017rainbow}, which combines DQN \citep{mnih2015human} with double Q-learning~\citep{van2016deep,hasselt2010double}, prioritised replay~\citep{schaul2015prioritized}, dueling networks~\citep{wang2015dueling}, multi-step targets~\citep{sutton1988learning,Sutton:2018}, distributional RL~\citep{bellemare2017distributional}, and parameter noise for exploration~\citep{fortunato2017noisy}. Rainbow obtains median human-normalised score of 153\% on the human starts protocol and 223\% on the no-ops protocol. In contrast, the meta-gradient agent achieved a median score of 292.9\% on human starts and 287.6\% on no-ops, with the same number (200M) of frames. We note, however, that there are many differences between the two algorithms, including the deeper neural network architecture used in our work.

\section{Related Work}

Among the earliest studies on meta learning (or learning to learn~\citep{thrun1998learning}), \cite{schmidhuber1987evolutionary} applied genetic programming to itself to evolve better genetic programming algorithms. \cite{hochreiter2001learning} used recurrent neural networks like Long Short-Term Memory (LSTM)~\citep{hochreiter1997long} as meta-learners. A recent direction of research has been to meta-learn an \emph{optimiser} using a recurrent parameterisation~\citep{andrychowicz2016learning,wichrowska2017learned}. 
\cite{duan2016rl} and \cite{wang2016learning} proposed to learn a recurrent meta-policy that itself learns to solve the reinforcement learning problem, so that the recurrent policy can generalise into new tasks faster than learning the policy from scratch. Model-Agnostic Meta-Learning (MAML)~\citep{finn2017model,finn2017meta,finn2017one,grant2018recasting,al2017continuous} learns a good initialisation of the model that can adapt quickly to other tasks within a few gradient update steps. These works focus on a multi-task setting in which meta-learning takes place on a distribution of training tasks, to facilitate fast adaptation on an unseen test task. In contrast, our work emphasises the (arguably) more fundamental problem of meta-learning within a single task. In other words we return to the standard formulation of RL as maximising rewards during a single lifetime of interactions with an environment. 

Contemporaneously with our own work, \citet{zheng2018learning} also propose a similar algorithm to learn meta-parameters of the return: in their case an auxiliary reward function that is added to the external rewards. They do not condition their value function or policy, and reuse the same samples for both the update function and the cross-validation step -- which may be problematic in stochastic domains when the noise these updates becomes highly correlated.

There are many works focusing on adapting learning rate through gradient-based methods~\citep{sutton1992adapting,schraudolph1999local,maclaurin2015gradient,pedregosa2016hyperparameter,franceschi2017forward}, Bayesian optimisation methods~\citep{snoek2012practical}, or evolution based hyper-parameter tuning~\citep{jaderberg2017population, Elfwing:2017}. In particular, \citet{sutton1992adapting}, introduced the idea of online cross-validation; however, this method was limited in scope to adapting the learning rate for linear updates in supervised learning (later extended to non-linear updates by \citet{schraudolph1999local}); whereas we focus on the fundamental problem of reinforcement learning, i.e., adapting the return function to maximise the proxy returns we can achieve from the environment.  

There has also been significant prior work on automatically adapting the bootstrapping parameter $\lambda$. \citet{singh1998analytical} empirically analyse the effect of $\lambda$ in terms of bias, variance and MSE. \citet{kearns2000bias} derive upper bounds on the error of temporal-difference algorithms, and use these bounds to derive schedules for $\lambda$. \citet{downey2010temporal} introduced a Bayesian model averaging approach to scheduling $\lambda$. \citet{konidaris2011td_gamma} derive a maximum-likelihood estimator, TD($\gamma$), that weights the $n$-step returns according to the discount factor, leading to a parameter-free algorithm for temporal-difference learning with linear function approximation. \citet{white2016greedy} introduce an algorithm that explicitly estimates the bias and variance, and greedily adapts $\lambda$ to locally minimise the MSE of the $\lambda$-return. Unlike our meta-gradient approach, these prior approaches exploit i.i.d. assumptions on the trajectory of experience that are not realistic in many applications.

\section{Conclusion}

In this work, we discussed how to learn the meta-parameters of a return function. Our meta-learning algorithm runs online, while interacting with a single environment, and successfully adapts the return to produce better performance.  We demonstrated, by adjusting the meta-parameters of a state-of-the-art deep learning algorithm, that we could achieve much higher performance than previously observed on 57 Atari 2600 games from the Arcade Learning Environment.

Our proposed method is more general, and can be applied not just to the discount factor or bootstrapping parameter, but also to other components of the return, and even more generally to the learning update itself. Hyper-parameter tuning has been a thorn in the side of reinforcement learning research for several decades. Our hope is that this approach will allow agents to automatically tune their own  hyper-parameters, by exposing them as meta-parameters of the learning update. This may also result in better performance because the parameters can change over time and adapt to novel environments.

\section*{Acknowledgements}
The authors would like to thank Matteo Hessel, Lasse Espeholt, Hubert Soyer, Dan Horgan, Aedan Pope and Tim Harley for their kind engineering support; and Joseph Modayil, Andre Barreto for their suggestions and comments on an early version of the paper. 

\newpage
\bibliographystyle{abbrvnat}

\bibliography{nips_2018}

\newpage
\appendix 
\section{Detailed Hyper-Parameters used in the Atari Experiments}\label{appendix:hyperparam}
In Table~\ref{tb:hyperparams}, we describe the details of the important hyper-parameters used in the Atari experiments. The IMPALA hyper-parameter section is following~\citet{espeholt2018impala}, which is provided here for self-contained purpose. The hyper-parameters in meta-gradient section are obtained by a search on six games (Beamrider, Breakout, Pong, Q*bert, Seaquest and Space Invaders) following common practice in Deep RL Atari experiments~\citep{van2016deep, mnih2016asynchronous, wang2015dueling}. All of the hyper-parameters are fixed across all Atari games.
\begin{table}[h] 
\begin{center}
\begin{tabular}{l@{\hspace{5em}}l@{\hspace{.22cm}}}
\toprule
\textbf{IMPALA hyper-parameter} & \textbf{Value}  \\
\midrule
Network architecture & Deep ResNet \\
Unroll length & 20 \\
Batch size & 32 \\
Baseline loss scaling ($c$) & 0.5 \\
Entropy cost ($d$) & 0.01 \\
Learning rate ($\alpha$) & 0.0006 \\
RMSProp momentum & 0.0 \\
RMSProp decay & 0.99 \\
RMSProp $\epsilon$ & 0.1 \\
Clip global gradient norm & 40.0 \\
Learning rate schedule & Anneal linearly to 0 \\
                       & from beginning to end of training. \\
Number of learners & 1 (NVIDIA P100) \\
Number of actors & 80 \\
\hline\hline
\textbf{Meta-gradient hyper-parameter} & \textbf{Value}  \\
Trace decay ($\mu$) & 0 \\
Meta learning rate ($\beta$) & 0.001 \\
Meta optimiser & ADAM~\citep{kingma2014adam} \\
Meta batch size & 8 \\
Meta update frequency & along with every agent update \\
Embedding size for $\vet$  & 16\\
\hline
\end{tabular}
\end{center}
\caption{Detailed hyper-parameters for Atari experiments.}
\label{tb:hyperparams}
\end{table}

\section{Implementation Details} \label{appendix:impl}

\subsection{V-trace Return} \label{appendix:vtrace}
The $\lambda$-return \citep{sutton1988learning} is defined as
\[
G^{\lambda}_t = R_{t+1} + \gamma_{t+1} (1 - \lambda_{t+1}) v(S_{t+1}) + \gamma_{t+1} \lambda_{t+1} G^{\lambda}_{t+1} \,.
\]
This can be rewritten \citep{Sutton:2014} as
\begin{align*}
G^{\lambda}_t
& = v(S_t) + \delta_t + \gamma_{t+1} \lambda_{t+1} \delta_{t+1} + \ldots \\
& = v(S_t) + \sum_{k=0}^\infty \left(\prod_{j=1}^{k} \gamma_{t+j} \lambda_{t+j} \right) \delta_{t+k} ,
\end{align*}
where $\delta_{t} = R_{t+1} + \gamma_{t+1} v(S_{t+1}) - v(S_t)$, where we use the convention that $\prod_{j=1}^0 \cdot = 1$.

This return is on-policy.  For some algorithms, especially policy-gradient methods, it is important that we have an estimate for the current policy.  But in the IMPALA architecture, the data may be slightly stale before the learning algorithms consumes it.  Then, off-policy corrections can be applied to make the data on-policy again.  In particular, IMPALA uses a \emph{v-trace} return, defined by
\begin{align*}
G^{\lambda}_t
& = v(S_t) + \sum_{k=0}^\infty c_{t+k} \left(\prod_{j=1}^{k} \gamma_{t+j} c_{t+j} \right) \delta_{t+k} \,,
\end{align*}
where $c_t = \min\left(1, \rho_t \right)$ and $\rho_t = \frac{\pi(A_t|S_t)}{\pi'(A_t|S_t)}$, where $\pi$ is the current policy and $\pi'$ is the (older) policy that was used to generate the data.  Note that this return can be interpreted as an adaptive-$\lambda$ return, with a fixed adaptation scheme that depends only on the off-policy nature of the trajectory.  A similar scheme was proposed by \citet{Mahmood:2017}.

\subsection{Calculate the Meta-Gradient with Auto-Diff} \label{appendix:autodiff}
An important fact to note in the proposed approach is that, the update rule for $\th \rightarrow \th'$ in first-order optimiser like Equation~(\ref{eq:sgd}) is linear and differentiable. In modern machine learning frameworks like TensorFlow~\citep{abadi2016tensorflow}, we can alternatively obtain the meta-gradient specified in Equation~(\ref{eq:meta_chain_rule}), by utilising the automatic differentiation functionality in the framework. The only requirement is to rewrite the update operations so that the agent update can allow the gradient to flow through, since the build-in update operations are typically not differentiable in the common implementations. 

\subsection{Data Efficiency} \label{appendix:data_efficiency}
In order to reduce the data we needed for meta learning, we can reuse the experiences for both agent training and meta learning. For example, we can use experiences $\tau$ for updating $\theta$ into $\theta'$, validate the performance of this update via evaluating $J'$ on experiences $\tau'$. Vice versa, we can swap the roles of $\tau$ and $\tau'$, then use experiences $\tau'$ for updating $\theta$, and validate the performance of this update via evaluating $J'$ on experiences $\tau$. In this way, the proposed algorithm does not require extra data other than the ones used to train the agent parameter $\theta$ to conduct the meta learning update to $\vet$. 

\subsection{Running Speed} \label{appendix:speed}
As for running speed, with one learner on NVIDIA P100 GPU and 80 actors on 80 CPU cores, our method runs around 13k environment steps/second, compared to around 20k environment steps/second of IMPALA baseline on the same hardware and software environments. We introduce around $35\%$ additional compute overhead from the meta-gradient updates, however with this minor overhead we can boost the performance significantly. We'd like to highlight that the total wall clock time for finishing 200 Million frames in each game is about 4 hours.

\subsection{IMPALA}

We used the IMPALA algorithm with the \emph{deep} architecture and the \emph{experts} mode of training. In this mode, a separate agent is trained on each environment (i.e. the standard RL setting), as opposed to a multi-task setting. Population-based training was not utilised by the IMPALA experts in ~\citet{espeholt2018impala}; we follow this convention. In principle, $\gamma$ and $\lambda$ could be exposed to population-based training (PBT) ~\citep{jaderberg2017population}, however, this would blow up the computation time by the size of the population (24 in ~\citet{espeholt2018impala}), which is beyond reach for typical  experiments; furthermore adaptation by PBT does not exploit the gradient and is therefore perhaps less likely scale to larger meta-parameterisations.

\section{Results of Grid Search of Discount Factor $\gamma$ on Atari Experiments} \label{appendix:grid_search_gamma}

We conduct simple grid search on the discount factor $\gamma$, i.e., let $\gamma = 0.99, 0.995, 0.998, 0.999$ respectively, and apply it in the IMPALA framework. The grid search of discount factor $\gamma$ is to find some good $\gamma'$ ($\vet' = \gamma'$ in this case) to be used in the meta-objective $J'(\tau', \th', \vet')$, so that the meta learning approach can have a good proxy to the true return to learn from.

\begin{table}[h]
\centering
\caption{Performance comparison of IMPALA baseline with different discount factor $\gamma$, all scores are human-normalised~\citep{mnih2015human,wang2015dueling,van2016deep}.}
\label{tb:gamma_sweep}
\begin{tabular}{c|c|c}
\hline
                  & Human starts & No-ops starts \\ \hline
Agents            &  median   &   median   \\ \hline
$\gamma = 0.99$~\citep{espeholt2018impala}                    & 144.4\%  &        191.8\%  \\ \hline
$\gamma = 0.995$                     & 211.9\%                      &   257.1\%  \\ \hline
$\gamma = 0.998$                     & 208.5\%                &   210.7\%   \\ \hline
$\gamma = 0.999$                 & 114.9\%                      &   153.0\%    \\ \hline
\end{tabular}
\end{table}

As we can see from Table~\ref{tb:gamma_sweep}, the discount factor $\gamma$ has huge impact on the agent performance.

\section{Additional Experiment Results}

\subsection{Relative Performance Improvement in Individual Games} \label{appendix:individual}
\begin{figure}[h]
\centering
\includegraphics[width=0.95\textwidth]{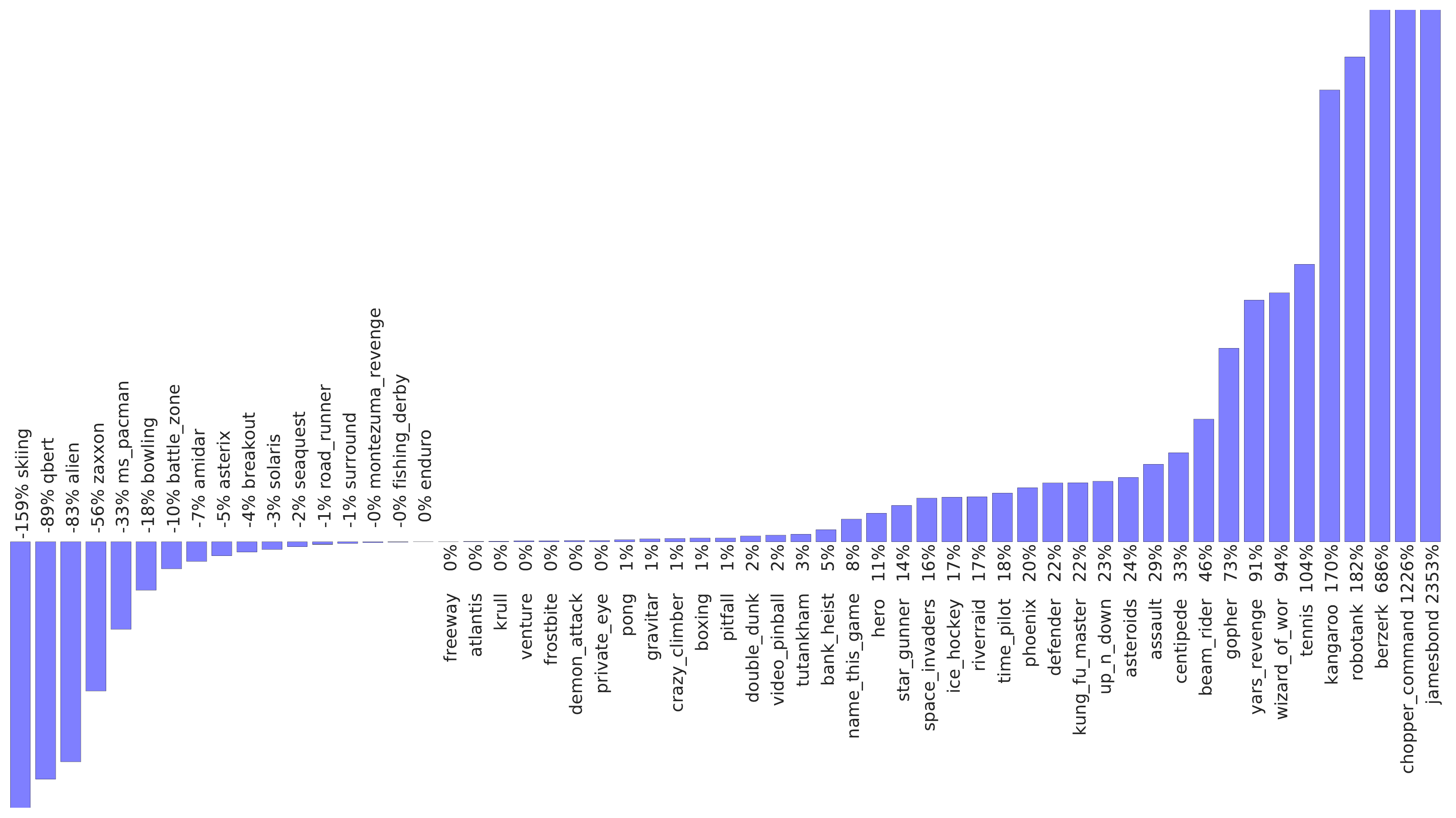}
\caption{The relative performance improvement of the meta-gradient algorithm, adapting discount factor, i.e., $\eta=\{\gamma\}$, compared to the baseline IMPALA ($\gamma = 0.99$) in  individual Atari 2600 games, where the gain is given by $\frac{\text{proposed} - \text{baseline}}{\text{max}(\text{human}, \text{baseline}) - \text{random}}$~\citep{wang2015dueling}. Improvement over 200\% is capped into 200\% for visualisation.
\label{fig:global_gamma_vs_baseline_99}
}
\end{figure}

\begin{figure}[h]
\centering
\includegraphics[width=0.95\textwidth]{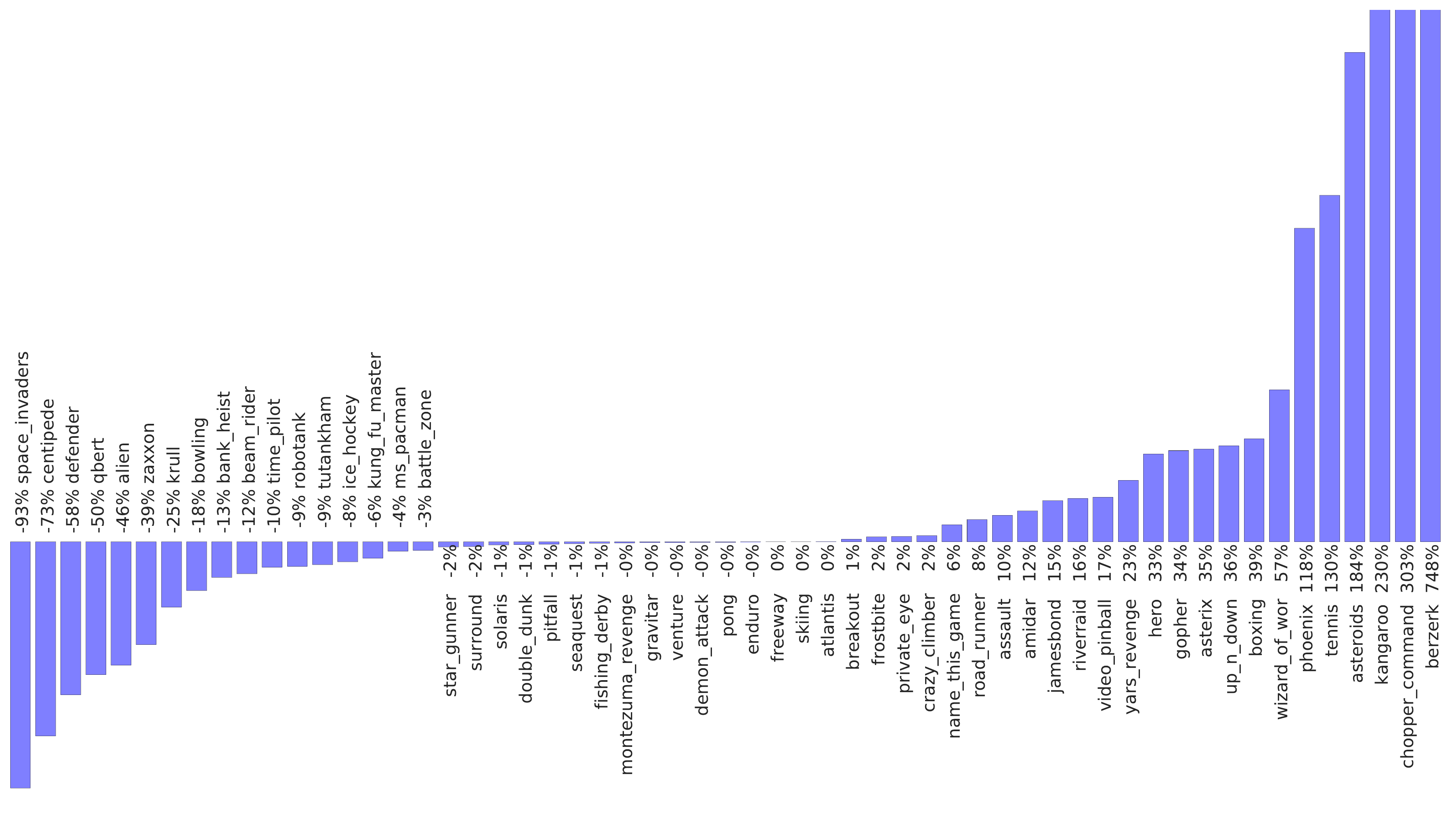}
\caption{The relative performance improvement of the meta-gradient algorithm, adapting both discount factor and bootstrapping parameter, $\eta=\{\gamma,\lambda\}$, compared to the baseline IMPALA ($\gamma = 0.995$) in  individual Atari 2600 games, where the gain is given by $\frac{\text{proposed} - \text{baseline}}{\text{max}(\text{human}, \text{baseline}) - \text{random}}$~\citep{wang2015dueling}. Improvement over 200\% is capped into 200\% for visualisation.
\label{fig:gamma_lambda_vs_baseline_995}
}
\end{figure}

In this section, we provide the relative performance improvement of the meta-gradient algorithm compared to the IMPALA baselines in individual Atari 2600 games. We show the results of adapting discount factor, i.e., $\eta = \{\gamma\}$, in Figure~\ref{fig:global_gamma_vs_baseline_99}, and results of adapting both discount factor and bootstrapping parameter, i.e., $\eta = \{\gamma, \lambda\}$, in Figure~\ref{fig:gamma_lambda_vs_baseline_995}.

\newpage
\subsection{Training Curves} \label{appendix:training_curves}
\begin{figure}[h]
    \centering
    game \hspace{1.0cm} $\gamma$ \hspace{1.2cm} 
    game \hspace{1.0cm} $\gamma$ \hspace{1.2cm} 
    game \hspace{1.0cm} $\gamma$ \hspace{1.2cm} 
    game \hspace{1.0cm} $\gamma$ 
    \includegraphics[width=\textwidth]{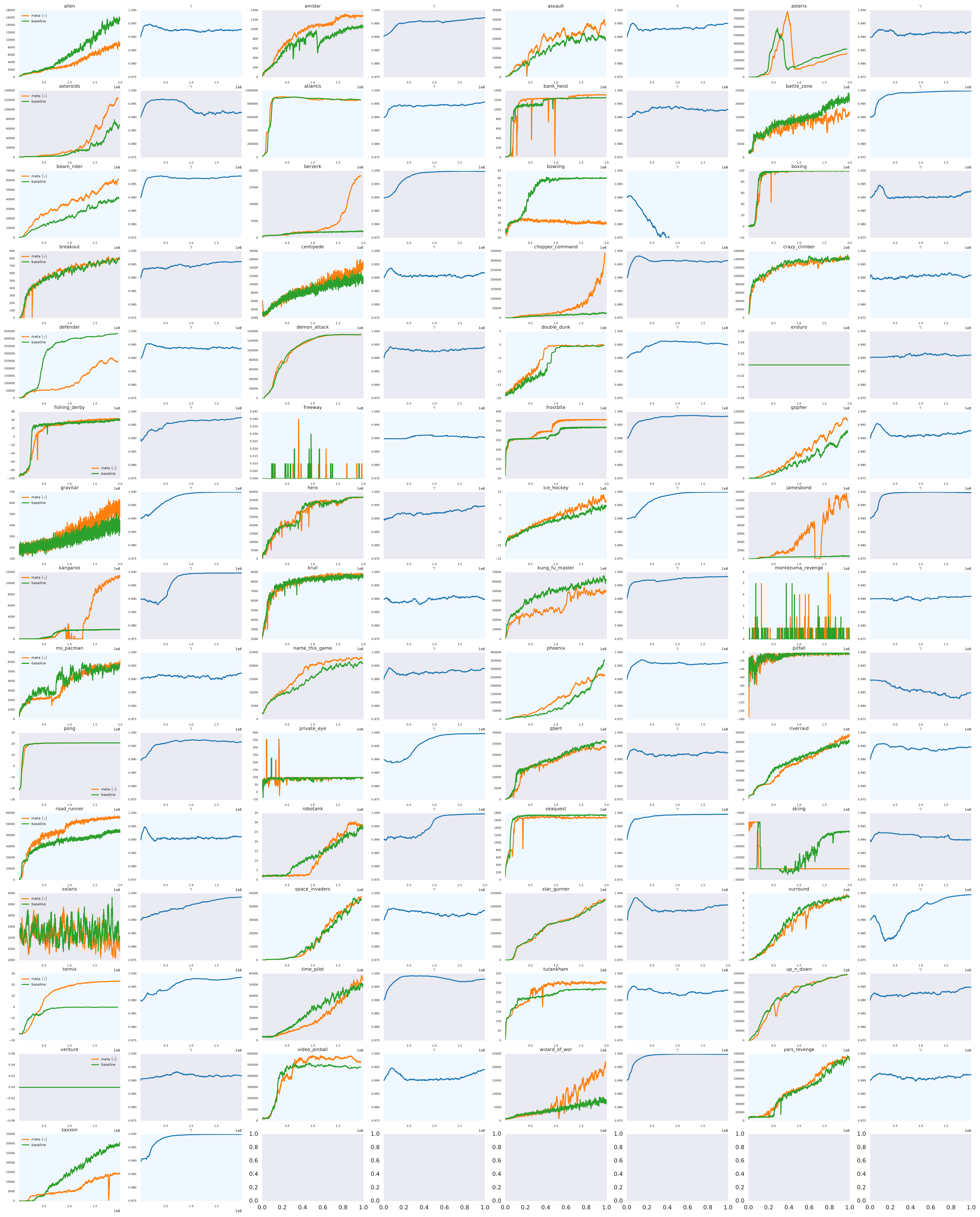}
    \caption{Training curves for meta learning $\vet = \{\gamma\}$ on $\gamma' = 0.99$. We provide the comparison of scores against baseline, and the change of $\gamma$ for each game. Best viewed in electronic version.}
    \label{fig:gamma_99_training_curves}
\end{figure}

\begin{figure}
    game \hspace{0.9cm} $\gamma$ \hspace{1.2cm} $\lambda$ \hspace{0.8cm}
    game \hspace{0.9cm} $\gamma$ \hspace{1.2cm} $\lambda$ \hspace{0.8cm}
    game \hspace{0.9cm} $\gamma$ \hspace{1.2cm} $\lambda$
    \centering
    \includegraphics[width=\textwidth]{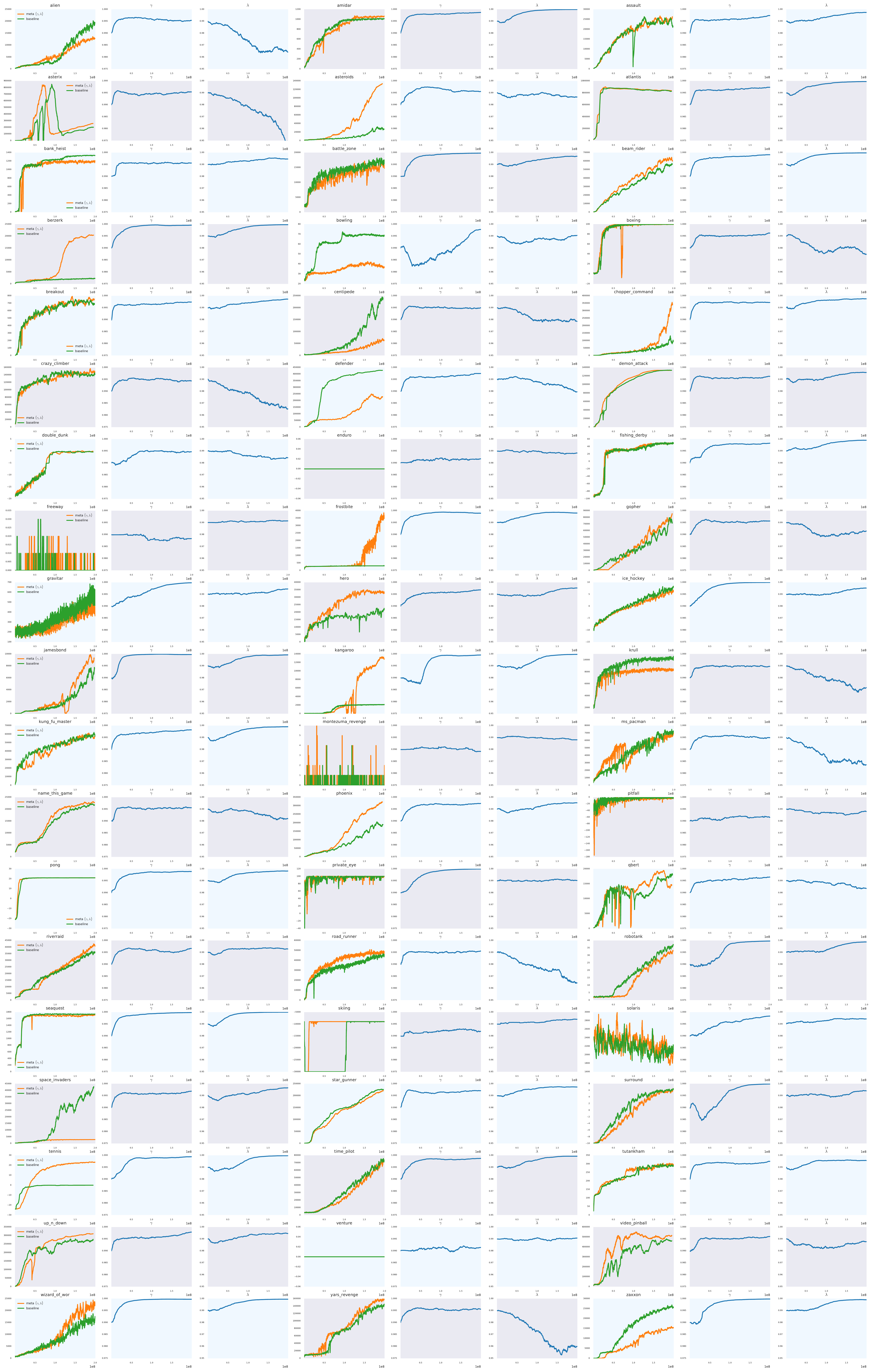}
    \caption{Training curves for meta learning $\vet = \{\gamma, \lambda\}$. We provide the comparison of scores against baseline, the change of $\gamma$, and the change of $\lambda$ for each game. Best viewed in electronic version.}
    \label{fig:gamma_lambda_training_curves}
\end{figure}

In this section, we provide the training curves for two representative experiments: adapting $\eta = \{\gamma\}$ with $\gamma'=0.99$ (Figure~\ref{fig:gamma_99_training_curves}) and adapting $\eta = \{\gamma, \lambda\}$ with $\gamma'=0.995$ (Figure~\ref{fig:gamma_lambda_training_curves}).

\end{document}